\documentclass[lettersize,conference]{IEEEtran}
\IEEEoverridecommandlockouts
\usepackage{cite}
\usepackage{amsmath,amssymb,amsfonts}
\usepackage{algorithm} 
\usepackage{algpseudocode} 
\usepackage{graphicx}
\usepackage{textcomp}
\usepackage{xcolor}
\usepackage{caption}% http://ctan.org/pkg/caption
\def\BibTeX{{\rm B\kern-.05em{\sc i\kern-.025em b}\kern-.08em
    T\kern-.1667em\lower.7ex\hbox{E}\kern-.125emX}}

\begin{document}

\title{Zero Touch Coordinated UAV Network Formation for 360$^\circ$ Views of a Moving Ground Target in Remote VR Applications
%UAV Based Tracking of a Moving Object Using Predator-Prey Swarm Interactions for 360 Views
}

\author{Yuhui~Wang and Junaid~Farooq\\
Department of Electrical and Computer Engineering, \\
University of Michigan-Dearborn, Dearborn, MI 48128 USA, \\
{Emails: \{ywangdq, mjfarooq\}@umich.edu}.
}

\maketitle

\begin{abstract}
Unmanned aerial vehicles (UAVs) with on-board cameras are widely used for remote surveillance and video capturing applications. In remote virtual reality (VR) applications, multiple UAVs can be used to capture different partially overlapping angles of the ground target, which can be stitched together to provide 360$^\circ$ views. This requires coordinated formation of UAVs that is adaptive to movements of the ground target. In this paper, we propose a joint UAV formation and tracking framework to capture 360$^\circ$ angles of the target. The proposed framework uses a zero touch approach for automated and adaptive reconfiguration of multiple UAVs in a coordinated manner without the need for human intervention. This is suited to both military and civilian applications. Simulation results demonstrate the convergence and configuration of the UAVs with arbitrary initial locations and orientations. The performance has been tested for various number of UAVs and different mobility patterns of the ground target.
\end{abstract}

%autonomously create formations that track the dynamic target on the ground as well as  for creating 


\IEEEpeerreviewmaketitle

%\vspace{-0.1in}
\section{Introduction}

%\textcolor{red}{the need of technology in the scenario: how and why (distributd, swarm, communication), challenges (in design of control framework, ibvs \& pred-pray), extend}
The virtual reality (VR) paradigm allows remote viewers to interact with a virtual environment in an immersive fashion~\cite{interconnected_VR}, and has been successfully applied in various domains such as %education \cite{vr_education}, 
remote monitoring \cite{remote_monitoring}, and target tracking \cite{vr_tracking}. In application scenarios like virtual tours, 
%traditional platforms \textcolor{red}{like work stations} may be limited by geographical or behavioral restrictions. However, 
unmanned aerial vehicles (UAVs) can be feasible platforms to deliver high fidelity real-time VR experiences for live viewers since they can capture videos of live scenery and transmit them to base stations (BSs) that service the VR users over wireless networks \cite{uav_vr}.
%However, traditional platforms like work stations may be limited by geographical or behavioral restrictions. 
%Unmanned aerial vehicles (UAVs) can be a feasible platform to deploy VR systems \cite{uav_vr} because they can capture videos on live games or scenery and transmit them to  base stations (BSs) that service the VR users over wireless networks.
%They are also used in first responder and emergency applications. 
In order to capture and transmit 360$^\circ$ views of a moving target on the ground, multiple UAVs with on-board cameras are required to be placed in the air such that partially overlapping images can be taken from all angles. Moreover, the UAVs need to maintain their relative positions alongside following the target when it moves.

The key objectives in creating such a formation using a swarm of UAVs are to (i) autonomously organize and place UAVs in a surround fashion without centralized planning and decision-making, (ii) continuously estimate the position of the ground target and track its motion (i.e., maintain a fixed distance with the target), and
(iii) center the target in the camera field-of- view (FOV) of all UAVs in real-time. These objectives are intertwined and a holistic approach needs to be developed to achieve the desired results. When 360$^\circ$ views of a moving target need to be captured and transmitted continuously, the estimation of the position and velocity of a moving object is challenging because the initial states of the object may be unknown, and measurements of the dynamics can fail when the object is occluded or moves out of view. Several approaches have been proposed for estimating the position or velocity of the target based on fixed camera \cite{fixed_cam}, sensor networks \cite{sensor_network} and radar \cite{radar}. Compared with traditional fixed-position platforms, UAVs have several advantages such as having a wide FOV, high mobility, and adaptability. However, to center the target in camera view, the design of the controller involves both the motion of the target and the UAV. If multiple UAVs are deployed, the swarm behaviors need to be adapted such that UAVs can captured images of the target from different perspectives. For effective stitching on multiple image views, the object need to be centered in each camera's point of view and cameras can capture images from diverse perspectives. 
%What is the problem? Keep the UAVs organized in a swarm for capturing different partially overlapping angles in a surround fashion, , and continuously track the ground object (maintain a fixed distance with the target)

%Why it is challenging to solve? 

\begin{figure}[t]
    \centering
    \includegraphics[width=3in]{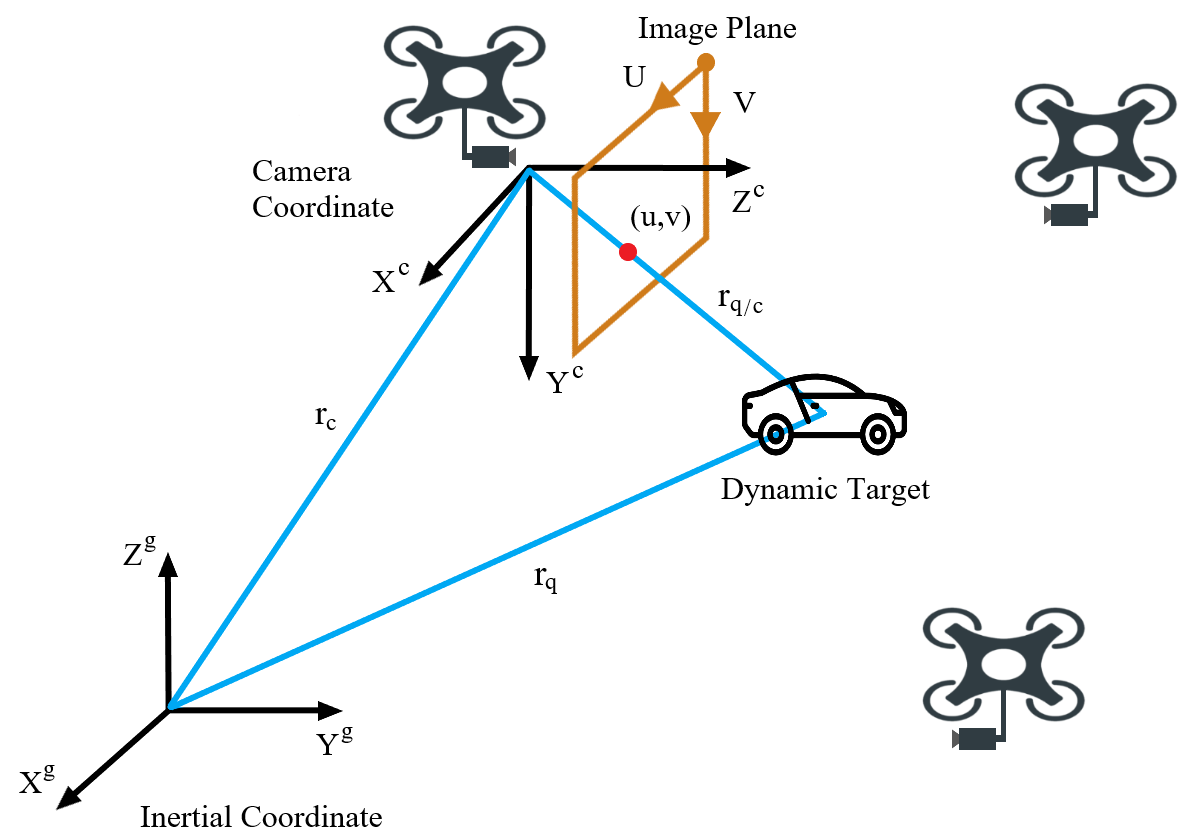}
    \caption{System architecture of a multi-UAV network formation for tracking and capturing 360 views of a dynamic target on the ground.}\vspace{-0.2in}
    \label{fig:fig1}
\end{figure}

In this work, we design and implement a UAV-based tracking algorithm inspired by the swarming interactions \cite{yuhui_icc_2022,yuhui_CNS_2022} to enhance formation resilience and track the target from multiple perspectives. An image-based visual servo (IBVS) \cite{ibvs} controller mechanism is adapted through the feedback dynamics of the UAVs to enable the target to be centered in all camera views. The framework is based on a zero touch philosophy, where the UAVs are able to configure themselves autonomously without human intervention and continuously adapt to achieve the desired group objective. The main contributions of the proposed framework are twofold: (i) a UAV motion controller based on the relative motion between UAVs and the target is designed to achieve the real-time tracking of the dynamic target, and (ii) a distributed and coordinated UAV formation and control algorithm is integrated to capture 360$^\circ$ views of the target.

%Inspired by the predator-prey interactions.

The rest of the paper is organized as follows: Section II provides an overview of the related works in literature, Section III describes the UAV kinematic model and camera model, Section IV details the proposed method for UAV formation control and object centering control, Section V provides the simulation results, and Section VI concludes the paper.

\section{Related Work}
Several works in literature focus on estimation and tracking of moving targets using cameras mounted on UAVs \cite{tracking_acc,uav_tracking_energy_efficient,uav_tracking_realtime,uav_tracking_recognition,vision_localization,uav_tracking_swarm,uav_tracking_learning,DL_360}. In \cite{tracking_acc,uav_tracking_energy_efficient}, the author proposed estimation and tracking algorithms of a moving target based on UAV cameras. They realized the challenges that motion pattern of the target is unknown and solve them through the use of unscented Kalman filter (UKF) \cite{ukf}. In their experimental results, a UAV with a mounted camera was implemented to capture image of the target and analyze the tracking path planning and energy consumption. But their proposed methods did not consider multiple UAV network formation and only worked in situations of single UAV tracking. The application can not provide full-view perspectives of the target and may lose track once the target is out-of-view of the UAV. 
Some researchers utilize static optimization or machine learning approaches for multi-UAV coordination \cite{uav_tracking_swarm,uav_tracking_learning}. Through the construction of a UAV system using computer vision algorithms like You Only Look Once (YOLO) \cite{yolo} and placement methods like reinforcement learning, their proposed methods can achieve both target tracking and UAV network formation. Although these works have achieved good performance in experimental results, they are limited by the centralized control architecture or unable to provide real-time decision-making.

Our proposed method is based on a distributed and coordinated UAV formation strategy which can create a uniform UAV network centered around the target. The designed controller considers both the motion of UAVs and the target, and utilizes images captured by the UAV-mounted camera to ensure that the target is always centered in all of the camera views.

\section{System Model}
Consider a set of UAVs $\mathcal{M} = \{1, \ldots, M\}$ tracking a dynamic target $\mathcal{Q}$. 
We assume that all UAVs are connected to a cellular base station for relaying the captured images and videos.
Each UAV can communicate with other UAVs that are within a distance $\gamma$. The set of neighboring UAVs of a UAV $i$ is defined by $\mathcal{N}_i$. For convenience of notation, we will drop the subscripts in the subsequent descriptions. For modeling simplicity, we assume that cameras are mounted on the UAVs without relative motion and we use a point model for the location of the UAVs and cameras by considering them to be pairwise co-located. The position, velocity and trajectory of the target is unknown but bounded and can be estimated by using a unscented Kalman filter~\cite{ukf}. Fig. 1 shows the relationship between a moving target and one of the cameras mounted on a UAV. The inertial coordinate is denoted by G and the camera coordinate is denoted by C. All coordinates are in inertial coordinate by default unless otherwise specified. The transformation matrix between coordinate systems G and C is provided below:

\begin{equation}
    \rm{T_{G2C}}=
    \begin{bmatrix}
        \rm{R}_{G2C} & p_c\\
        0 & 1
    \end{bmatrix},
\end{equation}
where the matrix $\rm{R}_{G2C}$ can be written as:
\begin{align}
    \begin{array}{ll}
    &\rm{R_{G2C}}=\rm{R}_x\cdot \rm{R}_y\cdot \rm{R}_z\\
    &=
    \begin{bmatrix}
        1 & 0 & 0\\
        0 & \cos(\theta_x) & \sin(\theta_x)\\
        0 & -\sin(\theta_x) & \cos(\theta_x)
    \end{bmatrix}\cdot
    \begin{bmatrix}
        \cos(\theta_y) & 0 & -\sin(\theta_y)\\
        0 & 1 & 0\\
        \sin(\theta_y) & 0 & \cos(\theta_y)
    \end{bmatrix}\cdot \\
    &\hspace{0.15 in}\begin{bmatrix}
        \cos(\theta_z) & \sin(\theta_z) & 0\\
        -\sin(\theta_z) & \cos(\theta_z) & 0\\
        0 & 0 & 1
    \end{bmatrix},
\end{array}
\end{align}

The parameters $\theta_x$, $\theta_y$, and $\theta_z$ are the rotation angles around $x$, $y$, $z$ axis respectively, $p_c=[x_c,  y_c,  z_c ]^T$ is the Cartesian coordinates of the UAV and can be obtained using global positioning system (GPS) measurements. 
The spatial velocity of the UAV is denoted by $V_c=[v_c,\omega_c]^T$, where $v_c,\omega_c\in\mathbb{R}^{3\times 1}$ are the linear and angular velocity of the camera respectively. The location coordinates of the target are denoted by $p_q=[x_q,  y_q,  z_q ]^T$ and need to be estimated. The spatial velocity of the target is denoted by $V_q=[v_q,\omega_q]^T$, where $v_q,\omega_q\in\mathbb{R}^{3\times 1}$ are the linear and angular velocity of the target respectively. We assume $\omega_q=[0,0,0]^T$ for the target vehicle. The relative position of the camera and the target can be computed as
\begin{equation}
    p_{q/c}=[X,  Y,  Z]^T=p_q-p_c,
\end{equation}
Using Corioli's theorem~\cite{Coriolis}, the relative velocity between the camera and the target can be written as
\begin{equation}
    \dot{p}_{q/c}=v_q-v_c-\omega_c \times p_{q/c},  
\end{equation}

where $v_c=[v_{cx},  v_{cy},  v_{cz} ]^T$ and $v_q=[v_{qx},  v_{qy},  v_{qz} ]^T$ are the linear velocity of camera and target respectively.

\subsection{Kinematics of the UAV}
As the measurement of positions of the target is through UAV-mounted cameras, a kinematic model \cite{vision_localization} based on the relative motion between the camera and the target is used to describe the change of system states over time. In \cite{tracking_acc}, the system states are designed as $X=[x_1,  x_2,  x_3,  x_q,  y_q,  z_q,  v_{qx},  v_{qy},  v_{qz}]^T$, where $[x_1,  x_2,  x_3 ]^T=[\frac{X}{Z},  \frac{Y}{Z},  \frac{1}{Z}]^T$ is defined to facilitate the matching with image pixel level coordinates. We assume the velocity of the target can be estimated by methods such as an unscented Kalman filter \cite{ukf} and is hence known to the UAVs. The dynamics of system states are given by

\begin{equation}
    \dot{X}=
    \begin{bmatrix}
        v_{qx} x_3-v_{qz} x_1 x_3+\zeta_1+\eta_1\\
        v_{qy} x_3-v_{qz} x_2 x_3+\zeta_2+\eta_2\\
        -v_{qz}x_3^2+v_{cz}x_3^2-(\omega_{cy}x_1-\omega_{cx}x_2)x_3\\
        v_q\\
        0\\
        0\\
        0
    \end{bmatrix},
    \label{eq:state_update}
\end{equation}
where $\zeta_1,\zeta_2,\eta_1,\eta_1 \in \mathbb{R}$ are defined by
\begin{equation}
    \begin{array}{cl}
        \zeta_1= & \omega_{cz}x_2-\omega_{cy}-\omega_{cy}x_1^2+\omega_{cx}x_1 x_2,\\
        \zeta_2= & -\omega_{cz}x_1+\omega_{cx}+\omega_{cx}x_2^2-\omega_{cy}x_1 x_2,\\
        \eta_1= & (v_{cz}x_1-v_{cx})x_3,\\
        \eta_2= & (v_{cz}x_2-v_{cy})x_3.
    \end{array}
\end{equation}

\subsection{UAV-mounted Camera Model}
Each UAV is equipped with an identical camera with focal lengths $f_x$ and  $f_y$ in pixel units. We assume that the relative position between the UAV and the camera is fixed and objects can be detected using computer vision algorithms like YOLO \cite{yolo}. The center of the the bounding box in the image plane can be determined and is denoted by $[u,  v]^T$. As $[u,  v]^T$ is also the projection point of the bounding box center on the camera plane, the relationship between system state and object position in camera view can be modeled as \cite{tracking_acc}

\begin{equation}
    x_1=\frac{u-c_u}{f_x},
\end{equation}
\begin{equation}
    x_2=\frac{v-c_v}{f_y},    
\end{equation}

where $[c_u,  c_v ]^T$ denote the center position of the camera view in pixel units.

\section{Methodology}
In this section, we present the methodology used to develop the UAV motion control.  In order to achieve both the coordinated formation as well as capturing fully centered views, two different controllers are adapted and integrated.

\subsection{Coordination UAV Formation Control}
The predator-prey swarm\cite{pred_prey} control is applied to maintain the autonomous and resilient formation of the UAV network. This controller enables the UAVs to form a centered formation around the target and keep the distance between UAVs at a desired value. The interaction function is defined by

\begin{equation}
    u_s=\sum_{i\in \mathcal{N}} k\left(\frac{r_j-r_i}{\|r_j-r_i\|^2}-\frac{r_j-r_i}{d_U^2}\right),
    \label{eq:eq_us}
\end{equation}

where $\mathcal{N}$ is the set of neighboring UAVs, $r_j$ is the position of UAV $j$ in inertial coordinates, $k$ is positive scaling constant and $d_U$ is the desired distance between UAVs. The advantage of this swarm control is that the distance based algorithm creates a UAV formation of a stable circle centered around the target with constant density. When the number of UAVs increase, the formation can automatically stratify into multiple layers and provide views from all perspectives.

\subsection{Object Centering Control}
The controller is designed based on both the swarm control and the IBVS control \cite{ibvs} which utilizes the camera images to achieve the tracking and centering of a dynamic target. The control state is defined by $s=[x_1,  x_2,  x_3]^T$ and the relationship between $\dot{s}$,$V_q$, and $V_c$, is given by
\begin{equation}
    \dot{s}=L_s(V_c-V_q),
    \label{eq:ls}
\end{equation}
where $L_s$ is the feature Jacobian matrix and is calculate by
\begin{equation}
    L_s=
    \begin{bmatrix}
        -x_3 & 0 & x_1x_3 & x_1x_2 & -(x_1^2+1) & x_2\\
        0 & -x_3 & x_2x_3 & x_2^2+1 & -x_1x_2 & -x_1\\
        0 & 0 & x_3^2 & x_2x_3 & -x_1x_3 & 0
    \end{bmatrix}.
\end{equation}
The state error is defined as the state deviation from the desired state
\begin{equation}
    e=s-s^*,
\end{equation}
where $s^*\in\mathbb{R}^3$ is the predefined desired state. The relationship between between $\dot{e}$, $V_q$, and $V_c$ is given by
\begin{equation}
    \dot{e}=L_e(V_c-V_q),
    \label{eq:le}
\end{equation}
where $L_e$ is the interaction matrix. As $\dot{e}=\dot{s}$, it can be shown that $L_e=L_s$ by comparing \eqref{eq:ls} and \eqref{eq:le}.
In order to locate the target at the center of camera view, its desired coordinates in camera plane are set as $[c_u,c_v]$. Then the desired state $s^*$ is calculated by
\begin{equation}
    s^*=\left[\frac{c_u-c_u}{f_x},\frac{c_v-c_v}{f_y},\frac{1}{d_q}\right]^T=\left[0,0,\frac{1}{d_q}\right],
\end{equation}
where $d_q$ is the desired distance between the camera and the target in inertial coordinates. In order to ensure an exponential decoupled decrease of error, we use an open-loop system
\begin{equation}
    \dot{e}=-\lambda e.
\label{eq:e2}
\end{equation}
Combining \eqref{eq:le} and \eqref{eq:e2} results in the following feedback control law 
\begin{equation}
    V_c =-\lambda L_e^+ e+V_q,
\end{equation}

where $L_e^+$ is the Moore-Penrose pseudo-inverse of $L_e$. With the pred-prey swarm control input $u_s$ in \eqref{eq:eq_us}, the final controller is designed as
\begin{equation}
    V_c =-\lambda L_e^+ e+V_q+u_s.
    \label{eq:update_vel}
\end{equation}

\subsection{Algorithm Design}
Based on the controllers discussed above, we proposed a UAV based tracking algorithm which provides 360$^\circ$ views of the target and enables distributed and autonomous formation of the UAV network as shown in Algorithm~\ref{alg:track_control}. Assume that the positions and velocities of UAVs are known, the positions and velocities of the target can be estimated \cite{ukf}. In every iteration, each UAV first updates its system state $X$ using the state dynamics in \eqref{eq:state_update}. Then the neighboring UAVs within the range $\gamma$ are identified. Each UAV shares its position and velocity with neighbors and finally updates its position and velocity using the discretized dynamics based on \eqref{eq:update_vel}. Our proposed distributed algorithm has linear time complexity of $\mathcal{O}(\mathcal{N})$, where $\mathcal{N}$ is the set of neighboring UAVs.

\begin{algorithm}
	\caption{Coordinated Formation and Tracking Control}
	\label{alg:track_control}
	\begin{algorithmic}[1]
	\Require Initialize position $p_q$ and velocity $V_q$ for the target. Initialize position $p_c$ and velocity $V_c$ for each UAV.
	\While {not converged}
	\For {each UAV}
	\State Estimate the velocity $V_q$ and position $p_q$ of the \\ \hspace{0.42 in}target.
	\State Update system states $X$ using \eqref{eq:state_update}.
	\State Determine the neighbors of the UAVs $\mathcal{N}$.
	\State Each UAV share its position and velocity with its \\ \hspace{0.42 in}neighbors.
	\State Update position $p_c$ and velocity $V_c$ using $\eqref{eq:update_vel}$.
	
	\EndFor
    \EndWhile
	\end{algorithmic} 
\end{algorithm}

\begin{figure}
    \centering
    \includegraphics[width = 3.5in]{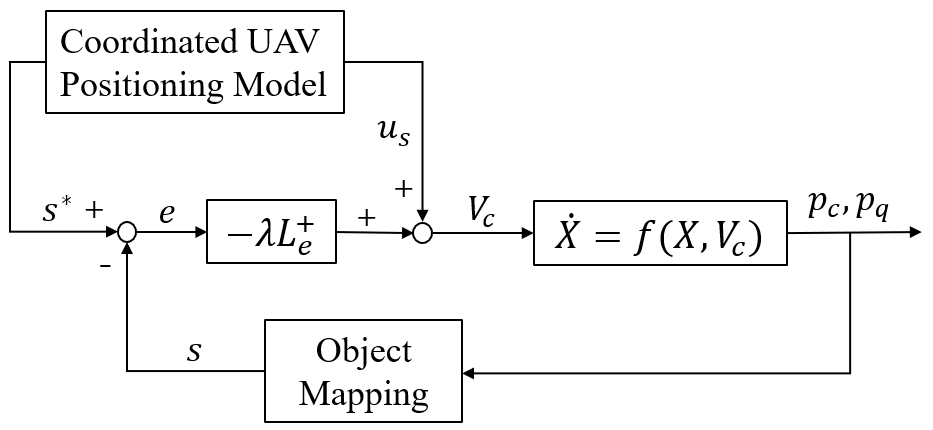}
    \caption{System model diagram.}\vspace{-0.1in}
    \label{fig:model_chart}
\end{figure}

\section{Performance Evaluation}
We conduct a series of simulations in MATLAB and measure the system performance in terms of state error, speed, and view angles. For the simplicity of implementation, we used a discretized system with the time step for every iteration $\Delta t=0.1s$.  The initial height of all UAVs are set to $50$ m, the initial position of the target is set to $[0,0,0]^T$, and the initial velocity of all UAVs are set to $[0,0,0]^T$. To test the tracking capability of the system, the velocity of the target is set to a continuously increasing trigonometric function with respect to time, i.e., $v_q(t)=[|t\sin(t)|,|t\cos(t)|,0]^T$. The parameters related to the system model are consistent during the simulation and are listed in table \ref{tab:table1}.

\begin{table}[h]
    \centering
    \begin{tabular}{|c|c|c|}
        \hline
        \textbf{Parameter} & \textbf{Symbol} & \textbf{Value} \\
        \hline
        Image Size &  & 640, 480 pixels\\
        \hline
        Image Center & $c_u,c_v$ & 320.5, 240.5 pixels\\
        \hline
        Focal Length & $f_x,f_y$ & 381.36, 381.36 pixels \\
        \hline
        Controller Parameters & $k,\lambda$ & 10.0, 1.0\\
        \hline
        Desired Distance & $d_U,d_q$ & 200m, 100m\\
        \hline
    \end{tabular}
    \caption{Parameter Settings}\vspace{-0.1in}
    \label{tab:table1}
\end{table}

\subsection{Performance Metrics}
To analyze the performance of the proposed framework, we define several metrics as follows: The \emph{normalized error} in $x,y,z$ coordinates of camera views is defined as $e_x=\frac{|u-c_u|}{c_u}$, $e_y=\frac{|v-c_v|}{c_v}$, $e_z=\frac{\|p_{q/c}\|_2}{d_q}$, and $e_x,e_y,e_z\in(0,1)$. The \emph{error area} is defined as $e_a=e_x*e_y$ and it can be used to quantify the deviation of the object from the image center. The \emph{total effective view angles} are defined to be the sum of non-overlapping perspective angles covered by all the cameras mounted on UAVs. The maximum view angle for each camera is set to 80 degrees for physical limitation. The total view angles of the system are calculated by the sum of view angles for all cameras and the effective view angles are the the non-overlapping views of all cameras. %To better visualize them, the total/effective view angles are normalized by 360 degrees.

\begin{figure}[t]
    \centering
    \includegraphics[width = 3.5in]{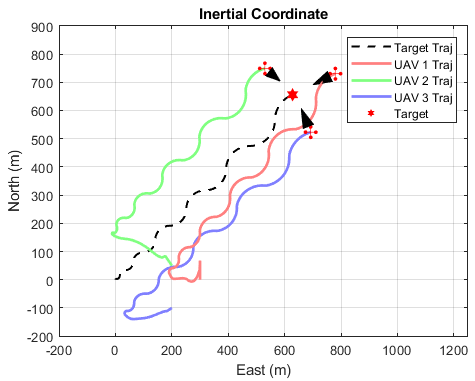}
    \caption{Top view of three UAVs tracking a moving target in inertial coordinates. Black arrows show the direction of UAVs. Trajectories in the first 200 iterations are shown.}
    \label{fig:result1}
    \vspace{-0.1 in}
\end{figure}

\begin{figure*}[t]
    \centering
    \includegraphics[width = \textwidth]{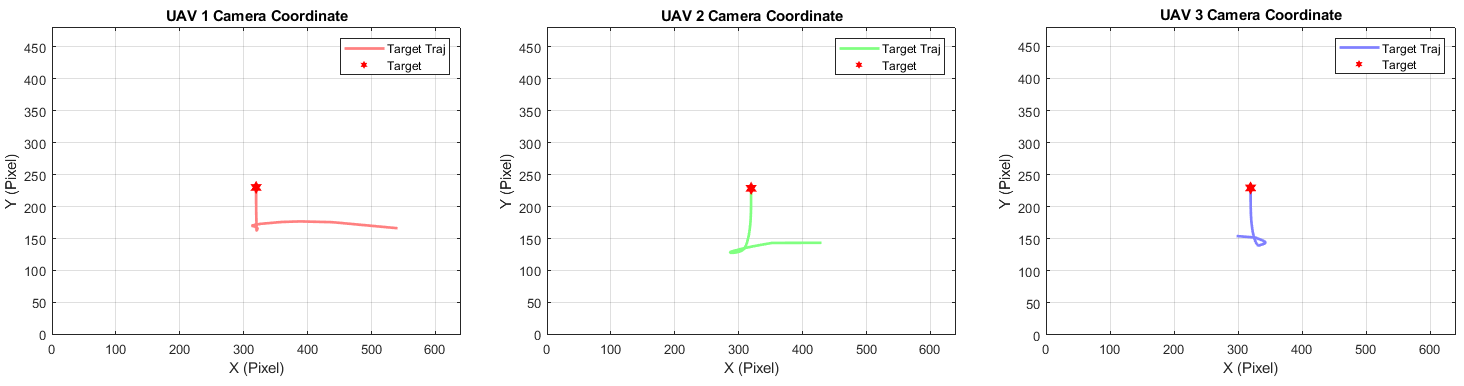}
    \caption{Camera views of three UAVs tracking a moving target in camera coordinate. The target is centered after convergence.}
    \label{fig:result2}
    \vspace{-0.0 in}
\end{figure*}

\subsection{Simulation Results}
\begin{figure}
    \centering
    \includegraphics[trim={10 5 20 20 in},clip,width = 3.5in]{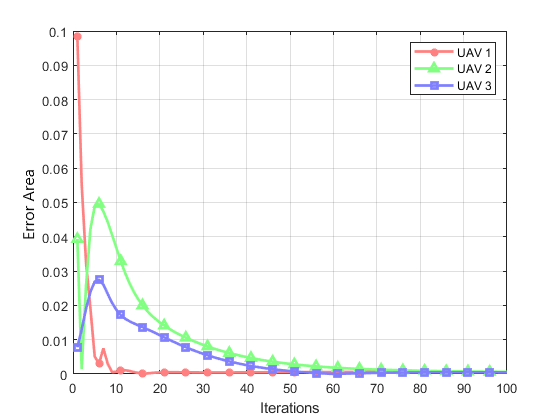}
    \caption{Evolution of error area over time for the UAVs. After initial glitches, the error diminishes over time.}
    \label{fig:result3}
    \vspace{-0.2 in}
\end{figure}

\begin{figure}
    \centering
    \includegraphics[trim={10 5 20 20 in},clip,width = 3.5in]{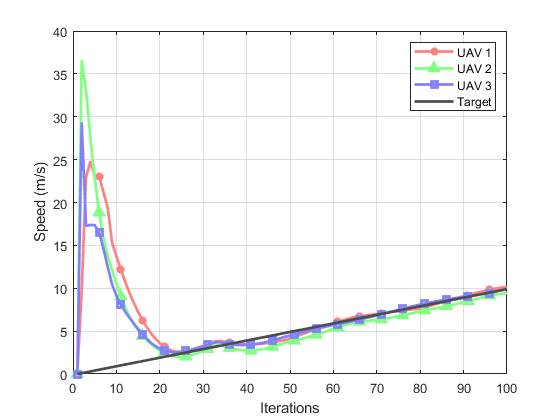}
    \caption{Comparison of the speed of the target and three UAVs in the first 100 iterations. The speed of the UAVs converge to the speed of the target after running for around 60 iterations.}
    \label{fig:result4}
    \vspace{-0.1 in}
\end{figure}

In the first simulation scenario, three UAVs were deployed to test the tracking capability and stability of the model. Fig. \ref{fig:result1} shows the top view results of the trajectories in inertial coordinate and the target trajectory in camera views were shown in Fig. \ref{fig:result2}. The initial positions of the UAVs were set to $[300,0,50]^T$, $[0,50,50]^T$, and $[0,-100,50]^T$ with the initial orientations $[0,0,-\frac{2}{3}\pi]^T,[0,0,-\frac{2}{3}\pi]^T,[0,0,\frac{5}{6}\pi]^T$ in radians. Black arrows represented the orientation of UAVs. The goal of the simulation was to ensure that the target was centered in camera views while keeping the distance between the UAV and the target at $d_q=100$ m, and the distance between UAVs at $d_U=200$ m. The area error rates over iterations were shown in the Fig. \ref{fig:result3} and the comparison between UAV speed and target speed was shown in Fig. \ref{fig:result4}. It showed that in the first 20 iterations, as the error $e_x$ in $x$ axis was large, the UAVs prioritized to adjust their positions so that the target moved to the center in camera views. At the same time, the swarm behavior controlled the UAVs to repel away from each other to maintain the inter-UAV distance at $d_U$. After 60 iterations, the UAV network converged to a stable formation and the UAVs could track the target and match its speed.

\begin{figure}[t!]
    \centering
    \includegraphics[trim={10 5 20 20 in},clip,width = 3.5in]{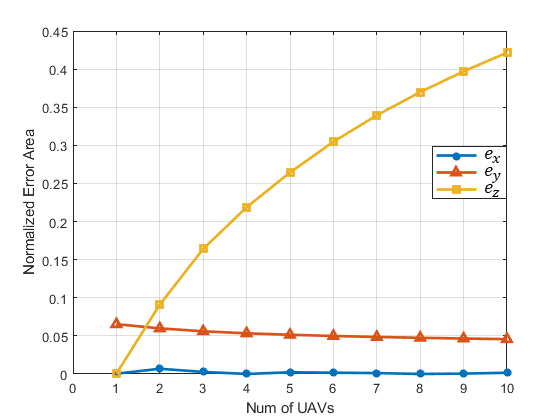}
    \caption{The normalized error in camera coordinates against number of UAVs.}
    \label{fig:result5}
    \vspace{-0.1 in}
\end{figure}

\begin{figure}[t!]
    \centering
    \includegraphics[trim={10 5 20 20 in},clip,width = 3.5in]{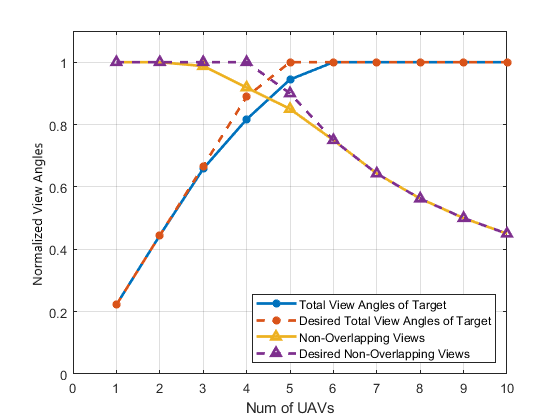}
    \caption{Comparisons between the desired/actual view angles and non-overlapping views against the number of UAVs.}
    \label{fig:result6}
    \vspace{-0.1in}
\end{figure}

In the second simulation, the performance of the system was tested as opposed to the number of UAVs. The UAVs ran for 200 steps and formed a centered formation around the target. Fig. \ref{fig:result5} showed the normalized error $e_x,e_y,e_z$ in $x,y,z$ axis respectively. The three curves demonstrated different behaviors of the system. The $e_x$ curve approached to the line $y=0$ because the $x$ coordinate of the target only depended on the orientation of the UAV and UAVs can rotate without physical limitations. As the number of UAVs increased, the $e_y$ curve approached to the line $y=0$ and the $e_z$ curve rose with decreasing slope. This was because as the number of UAVs increased, the radius of the formation also increased. $e_y$ was negatively related to the radius while $e_z$ was positively related to the radius. Fig. \ref{fig:result6} showed the actually achieved and optimal maximum view angles provided by the UAV network. The view angles were divide by $2\pi$ for normalization. The camera view angle of each UAV was 80 degrees. When the number of UAVs was less than 3, there was no overlapping views and the UAV formation was efficient. However, it cannot provide 360 degree seamless vision of the target. After the number of UAVs increased to 6, the UAVs can provide all-round views of the target. But there were overlapping views so the efficiency decreased. The comparison of the total and non-overlapping view angle curves revealed the trade-offs between efficiency and effectiveness.

\section{Conclusion}
In this paper, we have proposed an object tracking algorithm on UAV platforms with linear time complexity that can provide 360$^\circ$ views of a moving ground target and keep the autonomous and resilient formation of the UAV network. 
Both the positions and velocities of the target and UAVs are considered in the design of system states. Inspired from the IBVS and the predator-prey swarming approach, we designed an integrated system controller that minimize the deviation of the target feature point from the camera image center, keep a desired distance between the UAVs and the target, and achieve distributed physical orchestration of the UAV network. 
%Using the relative position between UAVs and the target, the proposed method enables the UAV to keep a desired distance with the target and keep the target in the center of the camera view. With the predator-prey swarm interaction, the UAVs can maintain the formation of the network and provide object tracking for seamless perspectives.
Simulation results showed that the target can be positioned to the center of all the camera views and UAVs can interactively adjust their positions centered at the target. After running for sufficient iterations, the errors converged to zero and the velocities of the UAVs matches the velocity of the target. The system performance can be observed to improve significantly as the number of UAVs increase. Our proposed method can well balance the tracking efficiency and effectiveness and be useful in constructing a zero touch coordinated UAV network, which can achieve seamless construction of 360$^\circ$ views suited for wireless VR applications.

\ifCLASSOPTIONcaptionsoff
  \newpage
\fi

\bibliographystyle{IEEEtran}
\bibliography{IEEEabrv,Bibliography}

\end{document}